\documentclass[runningheads]{llncs}
\usepackage{url}
\usepackage{comment}
\usepackage{amssymb}
\usepackage{booktabs} 
\usepackage{enumitem}
 \usepackage{array,multirow,graphicx}
 \usepackage{wrapfig,lipsum}
 \usepackage{float}
\usepackage{multirow}
\usepackage{graphicx}
\usepackage{enumitem}
\usepackage{pgfplots}
\usepackage{amsmath,amssymb,amsthm}
\usepackage{marvosym}

\newtheorem*{thm*}{Theorem}

\usepackage{mathtools}

\newcommand{\st}{\colon}

\newcommand{\img}[1]{\mathrm{Range}(#1)}

\renewcommand{\phi}{\varphi}

\newcommand{\dist}[2]{\mathrm{dist}(#1,#2)}

\newcommand{\set}[1]{\mathcal{#1}}
\newcommand{\Set}[1]{\set{#1}}

\usepackage{csquotes}
\usepackage{makecell}

\usepackage{tikz}

\usepackage{xspace}

\newcommand{\StdApproach}{DP-Embedding\xspace}
\newcommand{\EffApproach}{Personalized DP-Embedding\xspace}

\usepackage[symbol]{footmisc}

\usepackage{algorithm,algpseudocode}
\usepackage{algcompatible}
\usepackage{upgreek}
\usepackage{subcaption}
\usepackage{comment}
\usepackage{amssymb}
\newcommand{\Lili}[1]{\textcolor{red}{[Lili: #1]}} 
\newcommand{\LiliHighlight}[1]{\textcolor{blue}{[#1]}} 
\newcommand{\Sonvx}[1]{\textcolor{blue}{[XS: #1]}} 
\usepackage[normalem]{ulem} 
\newcommand{\SonT}[1]{\textcolor{green}{[SonTran: #1]}} 

\newcounter{Figcount}
\newcounter{tempFigure}
\newenvironment{figCaption}{%
	\renewcommand{\figurename}{Alg.}
	\setcounter{tempFigure}{\thefigure}
	\setcounter{figure}{\theFigcount}
}{%
	\setcounter{figure}{\thetempFigure}
	\stepcounter{Figcount}
}

\begin{document}

\title{dpUGC: Learn Differentially Private Representation for User Generated Contents}

\institute{}
\author{Xuan-Son Vu$^1$, Son N. Tran$^2$, Lili Jiang$^1$}
\authorrunning{Xuan-Son Vu, Lili Jiang}
\institute{$^1$Department of Computing Science, Ume\r{a} University, Sweden;\\
$^2$ICT Discipline, University of Tasmania, Australia;\\
\email{$^1$\{sonvx, lili.jiang\}@cs.umu.se,
	$^2$sn.tran@utas.edu.au
}
}

\authorrunning{Xuan-Son Vu et al.}
\titlerunning{dpUGC: Learn Differentially Private Representation for UGCs}

\maketitle

\begin{abstract}
\vspace{-25pt}
This paper firstly proposes a simple yet efficient generalized approach to apply differential privacy to text representation (i.e., word embedding). Based on it, we propose a user-level approach to learn personalized differentially private word embedding model on user generated contents (UGC).  To our best knowledge, this is the first work of learning user-level differentially private word embedding model from text  for sharing. The proposed approaches protect the privacy of the individual from re-identification, especially provide better trade-off of privacy and data utility on UGC data for sharing.  The experimental results show that the trained embedding models are applicable for the classic text analysis tasks (e.g., regression).  Moreover, the proposed approaches of learning differentially private embedding models are both framework- and data-independent, which facilitates the deployment and sharing. The source code is available at \url{https://github.com/sonvx/dpText}.
\end{abstract}


\section{Introduction}\label{introduction}

\begin{figure}[h]
	\renewcommand{\figurename}{Figure}
	\centering
	\includegraphics[height=0.326\textheight, width=0.60\textwidth]{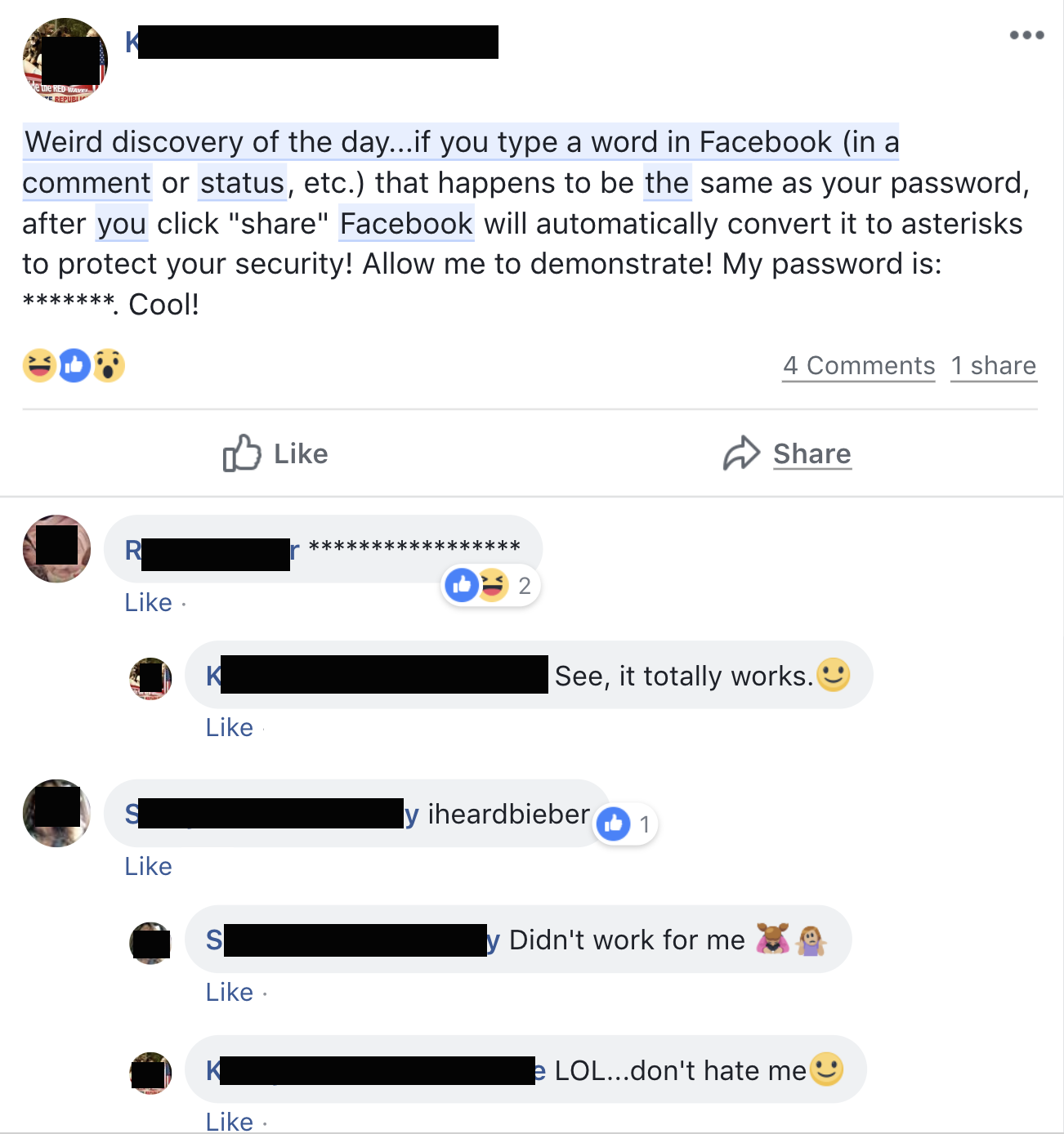}
    \vspace{-0.3cm}
\caption{A prank causes user credentials leak in FB data
}
\label{fig:fbprank}
    \vspace{-0.6cm}
\end{figure}

\begin{figure}[!ht]
\centering
\includegraphics[width=13cm]{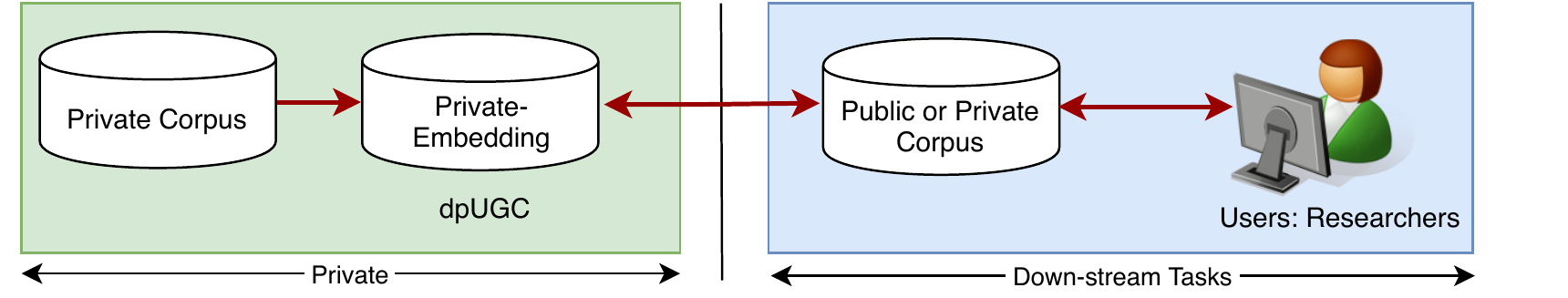}
\vspace{-0.6cm}
\caption{Overview of our safe-to-share embedding model that can be used to facilitate research on sensitive data with privacy-guarantee.}
\label{fig:privacy_schema}
\vspace{-0.6cm}
\end{figure}

Word embedding, also known as word representation, represents a word as a vector capturing both syntactic and semantic information, so that the words with similar meanings should have similar vectors~\cite{Levy:2014}.  This representation has two important advantages: efficient representation due to dimensionality reduction, and semantic contextual similarity due to a more expressive representation.
Thanks for these advantages, word embedding is widely used to learn text representation for text analysis tasks. Some commonly used word embedding models include Word2Vec~\cite{Mikolov:13}, GloVe~\cite{pennington2014glove}, and FastText~\cite{bojanowski2017enriching} and successfully applied in a variety of tasks like parsing~\cite{Bansal:2014}, topic modeling~\cite{Batmanghelich:2016}, and document classification~\cite{Taddy:2015}. Training word embedding model on big data requires high performance computing resources. For example, Word2Vec model was learned on 100 billion words from Google News corpus, and the FastText model of Facebook was learned from 840 billion words. Thus, once an efficient word embedding model was trained, it is most likely to be widely shared among researchers and communities.  However, since word embedding models preserve pretty much semantic relations between words, the shared pre-trained models may lead to privacy breaches especially when they were trained from UGC data such as tweets and Facebook posts.  For instance, user \emph{first name} (e.g., ``John''), \emph{last name} (``Smith'') and \emph{disease} (e.g., ``prostatitis'') may be represented as similar vectors in word embedding model.  Even user real name is absent from the pre-trained models, other available information such as \emph{username, address, city name, occupation}, could be represented with similar vectors, with/without auxiliary data, leading to re-identification risk to discover the individual to which the data belongs to, by using some approaches like author identification \cite{Mohsen:2016}, age and gender prediction~\cite{FlekovaG13}. Even further, the latent privacy breaches may cause a follow-up security issue.  Figure~\ref{fig:fbprank} shows a prank on Facebook to get other users' passwords. In case this type of information is learned and embedded in the embedding model, there exists a possible risk that one can exploit \emph{user} as a query to the shared embedding model and get their \emph{password}. One might argue that the sensitive information likes \emph{user, password} should not be leaked out and should have been removed from the embedding model. However, the purpose of learning from sensitive data is to learn the model without privacy leakage for facilitating research on sensitive data. To protect privacy, we statistically guarantee the chance to re-identify individuals by using output from the pre-trained models. Thanks to that, further research on the sensitive data \textbf{at large scale} can be possible such as ``what is the common patterns between users when they configure their passwords?" (to analyze security risks) or ``what diseases are normally unspeakable but get shared online?" (to analyze user behaviours on social networks). 
Similarly, this approach can be applied to user-level medical text data, which is very sensitive, to make research on medical data possible.
Figure~\ref{fig:privacy_schema} shows our approach to learn data distribution from private UGC data to facilitate studies on down-stream tasks.

As discussed above, it is critical to protecting privacy when learning embedding model for UGC data sharing. To address the challenge of revealing information about an individual in the training data, Dwork et al.~\cite{Dwork:2006,Dwork:2009} proposed differential privacy technique which provides a strong guarantee of privacy, and soon became a well known standard in privacy preservation. 
However, differential privacy is a general mechanism and how to apply to different data type is a non-trivial problem.
Some previous work applying differential privacy on text data ~\cite{McMahan:2018,Zhang:2018,Popov:2018} was either dependent on pre-defined sensitive features or applicable on federated framework instead of centralized data. The main difference is that they applied differential privacy to prevent the text data from personal data breaches, while this paper applies differential privacy to learn a shareable word embedding model from text data. Therefore, more challenges on privacy-budget control and data utility preservation have to be addressed.


\subsection{Goal of the Paper}

The goal of this paper is to develop effective and efficient approaches to apply differential privacy on text data to learn differentially private word embedding models. The ultimate purpose is to share the trained word embedding model, which prevents the highly latent risk of privacy breaches in word embedding models learning from UGC data and meanwhile maintains reasonable data utility. The main contributions of this paper are: 

\vspace{-3pt}
\setlength\parindent{20pt}
\begin{itemize}[wide]
\item We propose a simple yet efficient generalized approach of applying differential privacy on text data to learn embedding model for UGC data sharing.
\item We apply user-level privacy-guarantee on above differentially private word embedding model to maintain better data utility.
\item We conduct extensive experiments to evaluate the effectiveness of our proposed approach to preserve data utility, especially we test the approaches on text analysis task (i.e., regression).
\end{itemize}

The rest of this paper is organized as follows. Subsection~\ref{relatedwork} presents related work. Section~\ref{preliminaries} shows some preliminaries of differential privacy and word embedding. Section~\ref{proposedmethod} presents the proposed approaches to learn differentially private word embedding. Experiments settings and evaluation results are discussed in Section~\ref{experiment_settings} and~\ref{evaluations}. Section~\ref{conclusions} concludes the paper followed by future work.

\subsection{Previous Work\label{relatedwork}}
Anonymization~\cite{Bayardo:2005} and sanitization~\cite{WANG:2009} have been widely used in privacy protection.  \textbf{Differential Privacy} later emerged as the key privacy guarantee by providing rigorous, statistical guarantees against any inference from an adversary \cite{Dwork:2006}. Differential privacy has been applied in many research on different types of data including images~\cite{Fredrikson:2015,Abadi:2016,Papernot:2018,Wu:2018}, network~\cite{Nguyen:2016}, text~\cite{McMahan:2018,Zhang:2018,Popov:2018}, and general neural network architectures~\cite{Phan:2017}.  There is a family of algorithms called Private Aggregation of Teacher Ensembles (PATE), which becomes popular and contributes to research on differential privacy for machine learning. The advantage of PATE is to achieve private learning by coordinating the activity and sharing weights between different learning models ~\cite{Papernot:2018}.  Two limitations of PATE include: 1) it lacks flexibility and modularity when integrating to other frameworks, and 2) PATE was only trained on image data, which is not applicable to text data. Due to the different representation formats, differential privacy on text data reserves more difficulties.  To apply differential privacy on text data, ~\cite{McMahan:2018,Zhang:2018,Popov:2018} transformed the problem of differential privacy on text to language modeling problem, which aims to protect privacy for next word suggestion task on user devices. They are different from what we are addressing in this paper since their federated models require both clients and servers. Whereas our differentially private word embedding model is learned from centralized data, and further used for sensitive data sharing.
\section{Preliminaries}
\label{preliminaries}

\subsection{Differential Privacy} \label{subsec:dp}

To address the challenge of revealing information about an individual in the training data, \textbf{differential privacy}~\cite{Dwork:2006,Dwork:2009,Lee:2011,Lee:2012} essentially hides any individual by ensuring that the resulting model is nearly indistinguishable from the one without that individual. Differential privacy provides a strong guarantee of privacy even when the adversary has arbitrary external knowledge.  The basic idea is to add enough noise to the outcome (e.g., the model resulting from training) to hide the contribution of any single individual to that outcome. Let $D$ be a collection of data records, and one record corresponds to an individual. A mechanism $\Set{M}: D\rightarrow \mathbb{R}^d$ is a randomized function mapping database $D$ to a probability distribution over some range.  $\Set{M}$ is said to be differentially private if adding or removing a single data record in $D$ only affects the probability of any outcome within a small multiplicative factor.  The formal definition of ($\epsilon, \delta$) differential privacy is:

\begin{definition}
\textbf{[($\epsilon$-$\delta$)-differential privacy]}
\label{defn:diff-privacy}
\em{A randomized mechanism $\Set{M}$ is ($\epsilon$, $\delta$)-differential privacy where $\epsilon\geq0, \delta\geq0$, if for all data records in $D$ and $D'$ differing on at most one record, and $\forall \Set{S}\subseteq \img{\Set{M}}$:
	$$\Pr\left[\Set{M}(D)\in\Set{S}\right]\leq e^{\epsilon}\times \Pr\left[\Set{M}(D')\in\Set{S}\right]+\delta\ $$}
\end{definition}
The values of $(\epsilon,\delta)$ here are called \textbf{privacy-budget}. They control the level of the privacy, i.e., smaller values of $(\epsilon,\delta)$ guarantee better privacy but lower data utility.

\textbf{Privacy-budget}: There are typically two types of privacy-budget: (1) global privacy-budget~\cite{PINQ:09}, and (2) personalized privacy-budget~\cite{Ebadi:2015}. The main difference is that, the global budget is counted to all users while the personalized budget is counted based on different users. Therefore, personalized privacy-budget is a better way to control data utilities and privacy due to the fine-grained privacy budget. 

As introduced above, $D$ and $D'$ are adjacent datasets differing on at most one record, McMahan et al.~\cite{McMahan:2018} introduced a \textit{user-level differential privacy}, where $D$ and $D'$ are adjacent datasets differing on at most one user's all records.   This definition will be used to form our \emph{Personalized DP-Embedding} algorithm in Section~\ref{proposedmethod}.

\subsection{Word Embedding}

Word embedding is one of the most popular representations of document vocabulary. Simply speaking, they are vector representations of particular words.  It is capable of capturing the context of a word in a document, semantic similarity, and relation with other words.  Word2Vec\cite{Mikolov:13} developed by Google, is one of the most popular technique to learn word embeddings using a shallow neural network. Specifically, they propose a neural network architecture (i.e., the skip-gram model) that consists of an input layer, a projection layer, and an output layer to predict nearby words. Given a sequence of words $w_1, \dots, w_T$ in a corpus, each word vector is trained to maximize the following log probability of neighboring words:

$$\frac{1}{T}\Sigma_{t=1}^{T}\Sigma_{j \in nb(t)} \text{ log } p(w_j|w_t)$$ where $nb(t)$ is the set of neighbouring words of word $w_t$ and $p(w_j|w_t)$ is the normalized exponential probability (i.e., hierarchical softmax) of the associated word vectors $\vec{w_j}$ and $\vec{w_t}$.

\section{Methodologies: Differentially Private Word Embedding}\label{proposedmethod}


This section describes our proposed approaches toward differentially private word embedding on text data. We start by introducing a generalized approach to learn differentially private word embedding (Subsection \ref{sec:dpEmbedding}). After that, we reformulate the word embedding learning problem to user-level for personalized differentially private embedding (Subsection \ref{sec:pdpEmbedding}).

\subsection{\textbf{Differentially Private (DP-) Embedding}} \label{sec:dpEmbedding}


Differently from most of the previous studies, which apply differential privacy to image data, we implement algorithms for learning differentially private word embedding on text data. Compared with image data (represented by pixel positions, sizes of geometric forms, shapes etc.), text data captures more semantic and ambiguity, which makes it harder to preserve both privacy-guarantee and data utilities. Before introducing our approach, we formulate the problem of learning word embedding on text data as follows.

\textbf{Word embedding learning}:
Given a document corpus $D = \{d_1, \dots, d_n\}$,
each document $d\in D$ contains a sequence of words $\{w_1, ..., w_{|m|}\}$ from a fixed dictionary $V$. We use distributed representation and map every word in $V$ to a $k$-dimensional vector. The goal is to learn an embedding function $f$ that outputs a fixed length $k$ embedding for every $w \in V$. $k$ is called embedding size and typically from 50 to 300 dimensions.

As explained in Section \ref{subsec:dp}, the basic idea of differential privacy is to inject noise to a model in order to make more difficulty of predictableness, thus more difficult for hackers to predict the actual inputs. However, direct injection of noise to a trained embedding model will deteriorate the model's quality.  Technically, it would be better to insert noise during the learning process as we can optimize both the performance of the model and its privacy by treating the noise as a constraint. In the case of word2vec model, we can learn a differentially private embedding matrix $W$ by using a differential privacy optimizer such as DP-SGD (differentially private stochastic gradient descent)~\cite{Abadi:2016}. In particular, we apply noise to ``gradient'' during the training. As shown in Algorithm 1-a,
at each training step, a single training lot $L$ is used (line 3). Each training lot might have several minibatch $B$. But to make it simple, we consider $L = B$ in this case. A lot
$L$ is a random set of training samples in $D$ with a predefined
lot size $|L|$. Afterward, we compute gradients (line 4-5) which then will be added with noise (line 6-7) and applied to the standard gradient descent method (line 8-9). At line 10, a privacy accountant is used to accumulate privacy spending during the training process to ensure privacy guarantee. In the following, we will give more details regarding loss function and privacy accountant. 


\textbf{Loss function}: \emph{cross entropy loss} is used as our loss function, which measures the probabilistic distance between the predicted probabilities $p$ and the true binary labels $y$.
In our case, using one-hot encoding, the true label $y_i$ is 1 only when $w_i$ is the output word; $y_i$ is 0 otherwise. The loss function $\mathcal{L}_\theta$ of the model with parameter config $\theta$ aims to minimize the cross entropy between the prediction and the ground truth, as lower cross entropy indicates high similarity between two distributions.

\begin{equation}
\label{eq:1}
\mathcal{L}_\theta = - \sum_{i=1}^V y_i \log p(w_i | w_I) = - \log p(w_O \vert w_I)
\end{equation}

In the skip-gram model, the embedding matrix $W$ and output matrix $W'$ are a collection of input vectors and context vectors, respectively. Given one word $w_i$, its embedding vector  $\vec{w_i}$ is one row of $W$. Correspondingly, its context (output) vector $\vec{w'_i}$ is a column of the output matrix $W'$. The final output layer applies softmax to compute the probability of predicting the output word $w_O$ given $w_I$, and therefore:

$$p(w_O \vert w_I) = \frac{\exp({\vec{w'_O}}^{\top} \vec{w_I})}{\sum_{i=1}^V \exp({\vec{w'_O}}^{\top} \vec{w_I})}$$

Apply above to Equation (\ref{eq:1}), we have new loss function:

$$\mathcal{L}_{\theta} 
= - \log \frac{\exp({\vec{w'_O}}^{\top}{\vec{w_I}})}{\sum_{i=1}^V \exp({\vec{w'_i}}^{\top}{\vec{w_I} })}
= - {\vec{w'_O}}^{\top}{\vec{w_I} } + \log \sum_{i=1}^V \exp({\vec{w'_i} }^{\top}{\vec{w_I}})$$

In above loss function, the complexity of computing $\triangledown \text{ log }p(w_O|w_I )$ is proportional to $V$, which is often large ($10^5$ to $10^7$ terms).  In what follows, we will reduce the training cost by using \emph{Negative Sampling (\textbf{NEG})} ~\cite{Mikolov:13}, which was employed to train Google word2vec model. \textit{NEG} focuses on learning high-quality word embedding rather than modeling the word distribution in natural language. \textit{NEG} loss approximates the binary classifier's output with sigmoid functions.  Given an input word $w_I$, the correct output word is known as $w$. In the meantime, we sample $M$ other words from the noise sample distribution $Q$, denoted as $\tilde{w}_1, \tilde{w}_2, \dots, \tilde{w}_M \sim Q$. We label the decision of the binary classifier as $d$, which can only take a binary value ($d=1$ for positive samples, $d=0$ for negative samples). Thus, the final \textit{NEG} loss function looks like:


\begin{equation}
\label{loss_func}
\mathcal{L}_\theta = - [ \log p(d=1 \vert w, w_I) +  \sum_{i=1}^M\mathbb{E}_{\tilde{w}_i \sim Q} \log p(d=0|\tilde{w}_i, w_I)]
\end{equation}



\textbf{Differentially private word embedding model training:} we employ DP-SGD~\cite{Abadi:2016} to train the model using back-propagation.  
At each step of the differentially private SGD (see Algorithm 1-a), we compute the gradient $\triangledown \theta f(\theta, x_i)$ for a random minibatch $B$ (we consider $L = B$ in this case). Then we clip the 2-norm of each gradient belonging to the minibatch and compute their average. In the final step, noise is added in order to protect privacy before taking a gradient descent step using this "noisy" gradient. For dataset $D$, mechanism $\Set{M}$ (explained in Section \ref{subsec:dp}) is then given by:

$$\Set{M}(D) = \Sigma_{i \in B} \tilde \triangledown(f (x_i)) + \Set{N}(0, C^2\sigma^2 \textbf{I})$$

where $\tilde \triangledown(f (x_i))$ denotes the gradients clipped with a constant $C > 0$. The clipping is important since it helps to control gradient exploding and vanishing problem~\cite{Goodfellow:2016}.
\textbf{Privacy-accountant}: 
privacy-accountant keeps track of the privacy spendings through the whole training procedure. It is an important part of differentially private SGD. We applied
an ``accountant" procedure 
that computes the privacy spendings at each access to the training data and the accountant accumulates the cost at each step.


\textbf{Thoughts}: our experimental studies proved that the proposed approach above is an efficient way to learn differentially private word embedding to guarantee privacy.
In this approach,
the privacy-budget $(\epsilon, \delta)$ is either a hyperparameter (i.e., must set before training) or must be accumulated after each training epoch. The privacy budget, therefore,
is dependent on the number of training epochs, as it introduces noise into ``gradients" of parameters in every training step~\cite{Phan:2017}.  It is observed that when there is a small privacy budget, only a small number of epochs can be used to train the model~\cite{Abadi:2016}. While when the number of training epochs needs to be large to guarantee the model accuracy, the above approach may potentially sacrifice a portion of model utility.  This observation motivates us to further improve the proposed approach above. 


One intuitive solution is to inject noise differently to each part of the training data (e.g., add more noise into features which are less
relevant to the model output, and vice-versa~\cite{Phan:2017}). However, in reality, it is not always easy to reason what words in UGC are more significant/sensitive than others. For instance, political opinion in Asia countries is a sensitive topic and even forbidden by law in some countries (e.g., China). Conversely, they are less sensitive in the USA. Based on the above observations and thoughts, we propose ~\textbf{Personalised DP-Embedding} to control privacy-budget based on user privacy concerns.

\subsection{\textbf{Personalised DP-Embedding}} \label{sec:pdpEmbedding}
To achieve differential privacy in learning word embedding, we need user-level privacy~\cite{McMahan:2018}. Thus, we reformulate the problem of word embedding to accommodate user-level privacy, given the fact that in UGC, the mapping from user to data is known.  


\textbf{User-level word embedding learning}: Given a collection of user-level data $\{D_{1}, \dots, D_{u}, \dots, D_{k}\}$ where each user-level data $D_{u}$ contains a number of documents about user $u$. Without loss of generality, we define that a data collection $\Set{D}$ contains a set of $n$ document $\{d_1, d_2, \dots, d_n\}$ and $D_{u}\subseteq D$  $(1\leq u\leq k)$.  Each document $d$ contains a sequence of $m$ words \{$w_1, \dots, w_{m}$\} from a fixed dictionary $V$.  The different part from the original formalization (in Section \ref{sec:dpEmbedding}) is that, each user-level data $D_u$ has its own privacy-budget $(\epsilon, \delta)_u$. We do not have to set a predefined budget before learning or redistribute noise based on features as shown in ~\cite{Phan:2017}. Alternatively, we learn and protect people privacy based on their needs ~\cite{Vu:2018b}. During the training process, if privacy-budget $(\epsilon, \delta)_{u}$ of user $u$ is used up, the user-level data $D_u$ will no-longer be used (see Algorithm 1-b). In the algorithm (line 3), we firstly have to get a list of valid user-level data $D$ (by checking list of valid user $\Set{U}$). Then $L$ samples are drawn with probability $L/K$ (line 4). At line 5, we get list of users $\Set{U}_{L_t}$ where the sampled examples were taken. After this, we compute gradient (line 6-7), add noise (line 8-9), and go descent (line 10-11). At line 12, we compute the current privacy spending using privacy-accountant API of~\cite{Abadi:2016}. From line 13 to line 17, we update privacy spending for all users 
$\Set{U}_{L_t}$,  who get involved in the training step $t$, by the mean of privacy spending at that training step $t$. Then we exclude any user that got out of privacy-budget from $\Set{U}$.
One of the challenges in this approach is to obtain user privacy concerns. Based on a previous work~\cite{Vu:2017,Vu:2018b}, we found that the privacy-budget can be predicted using a strong correlation between user personality and their privacy concerns. Thus, we employed the model~\cite{Vu:2018b} to decide privacy-budget of user-level data.  Though the model was tested on a derived data, it is sufficient to predict privacy concern degree for cold-start users (i.e., having no user-defined privacy concern degree).


%


\textbf{Personalized privacy-accountant}: Based on the privacy accountant of the \StdApproach algorithm, we implement a personalized privacy-accountant for the \EffApproach algorithm to control privacy-budget of user-level data. Privacy-accountant can be used to 1) predict privacy-budget of all users if the information is not available (mentioned above), and 2) keep track of privacy-budget of each user to decide whether or not to use their data. 

\textbf{Remarks}: The advantage of this \EffApproach algorithm is that the user privacy-concerns will not be violated since they are defined by either user or algorithm (in case of cold-start users). Traditionally, the differential privacy based algorithm was learned based on a predefined $(\epsilon, \delta)$-budget to protect user data. In this way, the user level of privacy concerns was not considered and satisfied. Therefore, the proposed personalized DP-Embedding approach addresses this problem to fulfill the user needs of privacy.


\begin{figure}[!t]
	\scalebox{0.84}{
	\begin{minipage}[t]{14.2cm}
		\vspace{2pt}
		\begin{algorithm}[H]
			\label{naive_and_effective_approach}
			\begin{algorithmic}[1]
				\label{algo:NAIVEs-COMM}
				\REQUIRE Examples $\{x_1, \dots, x_N\}$, loss function $\Set{L(\theta)}$, embed dimension $k$
				\ENSURE return optimized $\theta$ to calculate $W^{(k)}$ - a learned DP-Embedding.
				\STATEx // \textbf{Algorithm 1-a}: \textbf{\StdApproach}
				\STATE Initialize $\theta_0$ randomly
				\FORALL {round $t = 0, 1, 2, \dots, T$}
				\STATE Take a random sample $L_t$ with sampling probability $L_t/N$
				\STATE \textbf{Compute gradient}
				\STATE For each $i \in L_{t}$, compute $g_t(x_i) \leftarrow \triangledown_{\theta_0} \Set{L}(\theta_t,x_i) \quad$ // $\mathcal{L}$  is from \eqref{loss_func}
				\STATE \textbf{Add noise}
				\STATE $\tilde g_t \leftarrow \frac{1}{L}(\Sigma_i \tilde g_t(x_i) + \Set{N}(0, \sigma^2C^2\textbf{I})$
				\STATE \textbf{Descent}
				\STATE $\theta_{t + 1} \leftarrow \theta_t - \eta_t \tilde g_t$
				\STATE $\Set{M.}$accum\_priv\_spending$(z)$
				\ENDFOR
				\\\hrulefill
			\end{algorithmic}
			\begin{algorithmic}[1]
			    \label{algo:EFF-COMM}
				\REQUIRE Examples $\{x_1, \dots, x_N\}$, loss function $\Set{L(\theta)}$, embed dimension $k$
				\ENSURE return optimized $\theta$ to calculate $W^{(k)}$ - a learned DP-Embedding.
				\STATEx // \textbf{Algorithm 1-b}: \textbf{\EffApproach}
				\STATE Initialize $\theta_0$ randomly
				\FORALL {round $t = 0, 1, 2, \dots, T$}
				\STATE $K \leftarrow$ (get list of samples from valid users $\Set{U}$)
				\STATE Take a random sample $L_t \in K$ with sampling probability $L_t/K$. 
				\STATE $\Set{U}_{L_t} \leftarrow$ the set of users where the sample $L_t$ come from.
				\STATE \textbf{Compute gradient}
				\STATE For each $i \in L_{t}$, compute $g_t(x_i) \leftarrow \triangledown_{\theta_0} \Set{L}(\theta_t,x_i) \quad$ // $\mathcal{L}$  is from \eqref{loss_func}
				\STATE \textbf{Add noise}
				\STATE $\tilde g_t \leftarrow \frac{1}{L}(\Sigma_i \tilde g_t(x_i) + \Set{N}(0, \sigma^2C^2\textbf{I})$
				\STATE \textbf{Descent}
				\STATE $\theta_{t + 1} \leftarrow \theta_t - \eta_t \tilde g_t$
				\STATE $(\epsilon_t, \delta_t) = \Set{M.}$get\_priv\_spending$(z)$
				\STATE \textbf{Update privacy spending for each user}
				\FORALL{user $u \in \Set{U}_{L_t}$}
				\STATE $(\epsilon, \delta)_{u} \leftarrow (\epsilon, \delta)_{u} + \frac{(\epsilon_t, \delta_t)}{L}$
				\STATE If user $u$ gets out of privacy-budget:
				 $\Set{U} \leftarrow \Set{U} \setminus \{u\}$
				\ENDFOR
				\ENDFOR
			\end{algorithmic}
		\end{algorithm}
	\end{minipage}
	}
   \begin{figCaption}
      \caption{Algorithms of \StdApproach and \EffApproach. $\Set{M}$ is the privacy account API of Abadi et al.~\cite{Abadi:2016}. 
      }
      \end{figCaption}
    \label{alg:naive_and_effective_approach}
    \vspace{-0.5cm}
\end{figure}
\vspace{-0.2cm}

\section{Experimental Settings}
\label{experiment_settings}
\subsection{Evaluation criteria}
We test our learned word embedding models on two criteria: 1) \textbf{word similarity}: it is a standard measurement for evaluating word embedding models~\cite{Levy:2014}.  The purpose is to detect the changes in semantic space. More similarity means less change in semantic space, which proves better word embedding model. One simple example is calculating similarity between the query word ($\vec{woman}$) and the predicted word $\vec{queen}$ given an embedding model (trained on $\vec{king}$ and $\vec{man}$);  2)\textbf{data utilities}: the main purpose of developing DP-Embedding is to preserve privacy when sharing the model for other scholars, and especially preserve data utility to facilitate their research.  Therefore, we evaluate the data utility of our learned embedding models by applying them to a downstream task - regression (more in Section \ref{subsec:regression}).
 
\subsection{Datasets} \label{subsec:dataset}

Two datasets are used for experimental evaluation: Text8~\footnote{\url{http://mattmahoney.net/dc/textdata.html}} dataset and myPersonality.org
(myPer) dataset.  Text8 dataset is commonly used to evaluate the quality of embedding models trained in different manners (e.g., normal embedding versus differentially private embedding). myPer dataset was used for data utility evaluation because of two reasons. Firstly, myPer dataset contains both a public set with 250 users and a private set with more than 153K users. Since early 2018, the private part of the myPersonality data is no longer available for scholars to apply, 
therefore, it increases the need for sharing information from the private data with privacy-guarantee than ever. Secondly, it fulfills the scenario this paper addresses, where the sensitive data has to be shared somehow for research benefits, and the urgency to guarantee privacy for data sharing.  Lastly, myPer dataset was widely used\footnote{https://goo.gl/M8iQ6m}\cite{mypersonality:2015} and we can conduct an evaluation by comparing the performance with previous works. Table~\ref{tbl:simple_statistics} summaries some statistics of the two datasets.

\begin{table}[]
	\centering
	\caption{A simple statistics of the myPersonality dataset and Text8 corpus.}
	\label{tbl:simple_statistics}
	\begin{tabular}{|p{2.5cm}|l|l|l|l}
		\hline
		Dataset &  \#users & \#documents & \#words \\ \hline \hline
		myPer (private) & 153,727 &  22,043,394 & 416,862,367 \\ \hline
		myPer (public) & 250 & 9,917 & 144,616 \\ \hline
		Tex8 corpus & - & - & 17,005,207 \\ \hline
	\end{tabular}
 \vspace{-0.8cm}
\end{table}

\subsection{Experiment design} \label{subsec:experimentdesign}
Two sets of experiments are designed to prove the effectiveness of the proposed DP-Embedding models regarding semantic space and regression task.
\textbf{Changes in Semantic Space} is detected to prove the effectiveness of preserving semantic relations of our \emph{DP Embedding} in comparison with a standard embedding. We firstly train a standard implementation of Word2Vec embedding (we refer it as \emph{Gold model}) on Text8 corpus. Secondly, we used our proposed approach to train the following two different word embedding models (1) DP-Embedding using the~\StdApproach and (2) None-DP Embedding model (without privacy guarantee) on Text8 corpus. Regarding evaluation, we issue the same set of queries to the DP-Embedding and None-DP Embedding and compare them with the returned top words from  \emph{Gold model}. The None-DP Embedding model is needed since we need to have a comparable learning pattern to compare to the DP Embedding, i.e., having a clipping gradient function.  Regarding evaluation metric, we used $MAP$ (mean-average-precision) to calculate word similarity (more details in Section \ref{subsec:semanticChange}). $MAP$ is widely used in information retrieval to evaluate results based on the top $K$ returned results. Given a list of queries $Q$ and their correct answers, $MAP$ metric calculates the mean of the average precision scores for each query, $MAP = \frac{\Sigma_{q=1}^{Q}Avg P(q)}{Q}$. Here we apply two different types of  $MAP$ called $MAP$-Word and $MAP$-Char. The $MAP$-Word evaluates the top similar words at word level, and the $MAP$-Char evaluates at character level. The difference between them is that, at word-level, $MAP$-Word will only capture exact words in the top results. However, during the training process, some similar words are at the top too but the $MAP$-Word cannot capture this information (e.g., ``there'' and ``that''). Inversely, the $MAP$-Char can capture very nicely this information at character level (see table~\ref{tbl:top_sim_words} for example of $MAP$-Word and $MAP$-Char).



\textbf{Regression task:} This experiment is used to prove that our differentially private models can preserve good data utility when they are shared for other scholars to use in a downstream application (e.g., regression). Here the regression task is to predict the extrovert personality score of people from the myPer(public) dataset. Given 250 users and the ground truth of extrovert scores, we divided the data to 80\% for training and the other 20\% for testing. 
As we mentioned in Section \ref{subsec:dataset}, there are two parts in myPer dataset (i.e., public and private). We first set up a experiment - E(public), where trained the public available word embedding model (Word2Vec) from Google and a character embedding model~\footnote{\url{https://github.com/minimaxir/char-embeddings/}} for feature representation. Meanwhile, we set up another experiment - E(private), where we trained our \emph{DP-Embedding} and \emph{None-DP Embedding} on myPer(private).
Based on the hypothesis that a regression task $R$ on features extracted from both public and private dataset will perform better than that only on public dataset (i.e., $R_{E(Private) + E(Public)} \geq R_{E(public)}$.), we compare the regression performance based on model from E(public) with that from both E(public) and E(private). To prove above hypothesis by evaluation, we implemented the following regression methods:

\begin{itemize}[wide]
\item Baseline-SVR: it is a regression baseline using Support Vector Machine-Regression (SVR) method, where only E(public) is used for feature extraction.
\item Baseline-LR: it is a regression baseline using linear regression (LR), where only E(public) is used for feature extraction.
\item DP-(SVM and LR): it is similar to the above methods except both E(private) and E(public) are used for feature extraction. E(private) here is trained using~\StdApproach.
\item NoneDP-(SVM and LR): it is similar to the above methods except we used E(private) and E(public) for feature extraction. E(private) here is trained without differential privacy.
\end{itemize}

\section{Evaluation Results}
\label{evaluations}

\begin{table}[t]
\begin{subtable}{.5\linewidth}\centering
	\scalebox{0.91}{
		\begin{tabular}{l|l|l|l|l}
    \hline
    Query & Gold model          & DP-Embedding (top 4)                     & MAP (W,C)      & Topic    \\ \hline
    three & four:two:five:seven & zero:one:feeder:nine                     & (0, 3.814)     & Numbers  \\ \hline
    eight & seven:nine:six:four & cornerback:four:stockholders:zero        & (0.5, 0.1347)  & Numbers  \\ \hline
    they  & we:there:you:he     & morgan:century:contentious:ferroelectric & (0, 0.4237)    & Pronouns \\ \hline
    \end{tabular}
    }
\caption{Top 4 on DP-Embedding model}\label{tab:first}
\end{subtable}%

\begin{subtable}{.5\linewidth}\centering
    \scalebox{0.99}{
        \begin{tabular}{l|l|l|l|l}
        \hline
        Query & Gold model          & Non-DP Embedding (top 4)                 & MAP (W, C)     & Topic    \\ \hline
        three & four:two:five:seven & one:in:UNK:zero                          & (0, 0.1288)    & Numbers  \\ \hline
        eight & seven:nine:six:four & integrator:transfection:four:one         & (0.33, 0.3561) & Numbers  \\ \hline
        they  & we:there:you:he     & that:monorail:it:lesbian                 & (0, 0.2341)    & Pronouns \\ \hline
        \end{tabular}
    	}
\caption{Top 4 on Non-DP Embedding model}\label{tab:second}
\end{subtable}%
	
	\caption{Top similar words of DP-Embedding (a), and Non-DP Embedding (b) models given three queries ``three'', ``eight'', and ``they'' at 100K learning step. The second column shows the best results from the Gold model. \small{MAP(W,C)} denotes \small{(MAP-Word,MAP-Char)}.}
	\label{tbl:top_sim_words}
	\vspace{-0.01cm}
\end{table}

\begin{table*}[t]
	\caption{Regression performance on public embedding with and without privacy guarantee in comparison with not using public embedding. Evaluation score is RMSE.
		$\dag$ marks good checkpoints to publish the DP-Embedding model, and LS stands for Learning\_Step.}
	\vskip 0.1cm
	\centering
	{
		\scalebox{0.812}{
		\begin{tabular}{l|l|l|l|l|l|l|l}
		\hline
		\multirow{2}{*}{LS} & \multicolumn{3}{l|}{SVR} & \multicolumn{3}{l|}{LR} & \multirow{2}{*}{Privacy-Budget (0.125, $\delta$)} \\ \cline{2-7}
                  & Baseline-SVR       &  DP-SVR     & NoneDP-SVR      & Baseline-LR      & DP-LR      & NoneDP-LR      &                   \\ \hline
			\hline
			20             & 2.6563        & \textbf{1.7881} & 3.5942          & 1.2903       & \textbf{1.2616} & 1.2642          & 0.0184 $\dag$            \\ \hline
			200            & 2.6563        & 2.4983          & \textbf{2.0198} & 1.2903       & \textbf{1.2589} & 1.2717          & 0.0189             \\ \hline
			500            & 2.6563        & \textbf{2.7795} & 3.6231          & 1.2903       & \textbf{1.2514} & 1.2909          & 0.0197 $\dag$            \\ \hline
			1K             & 2.6563        & 3.2146          & \textbf{2.0206} & 1.2903       & \textbf{1.2611} & 1.262           & 0.0211             \\ \hline
			5K             & 2.6563        & 6.1596          & \textbf{2.7472} & 1.2903       & \textbf{1.2577} & 1.2642          & 0.0372             \\ \hline
			10K            & 2.6563        & \textbf{1.6396} & 3.9155          & 1.2903       & 1.2768          & \textbf{1.2574} & 0.0755             \\ \hline
			50K            & 2.6563        & 2.9438          & \textbf{2.5769} & 1.2903       & 1.2574          & \textbf{1.2556} & 0.5929             \\ \hline
			90K            & 2.6563        & \textbf{2.4033} & 2.5175          & 1.2903       & 1.2585          & \textbf{1.258}  & 0.7681             \\ \hline
			100K           & 2.6563        & 2.6043          & \textbf{2.0215} & 1.2903       & \textbf{1.2548} & 1.262           & 0.7926             \\ \hline
		\end{tabular}
		} 
	}
	\label{tbl:linear_regression_results}
	\vspace{-0.2cm}
\end{table*}

\subsection{Evaluation \#1: Changes in Semantic Space} \label{subsec:semanticChange}

\begin{figure}
	\begin{subfigure}{.5\textwidth}
		\centering
		\includegraphics[width=.99\linewidth]{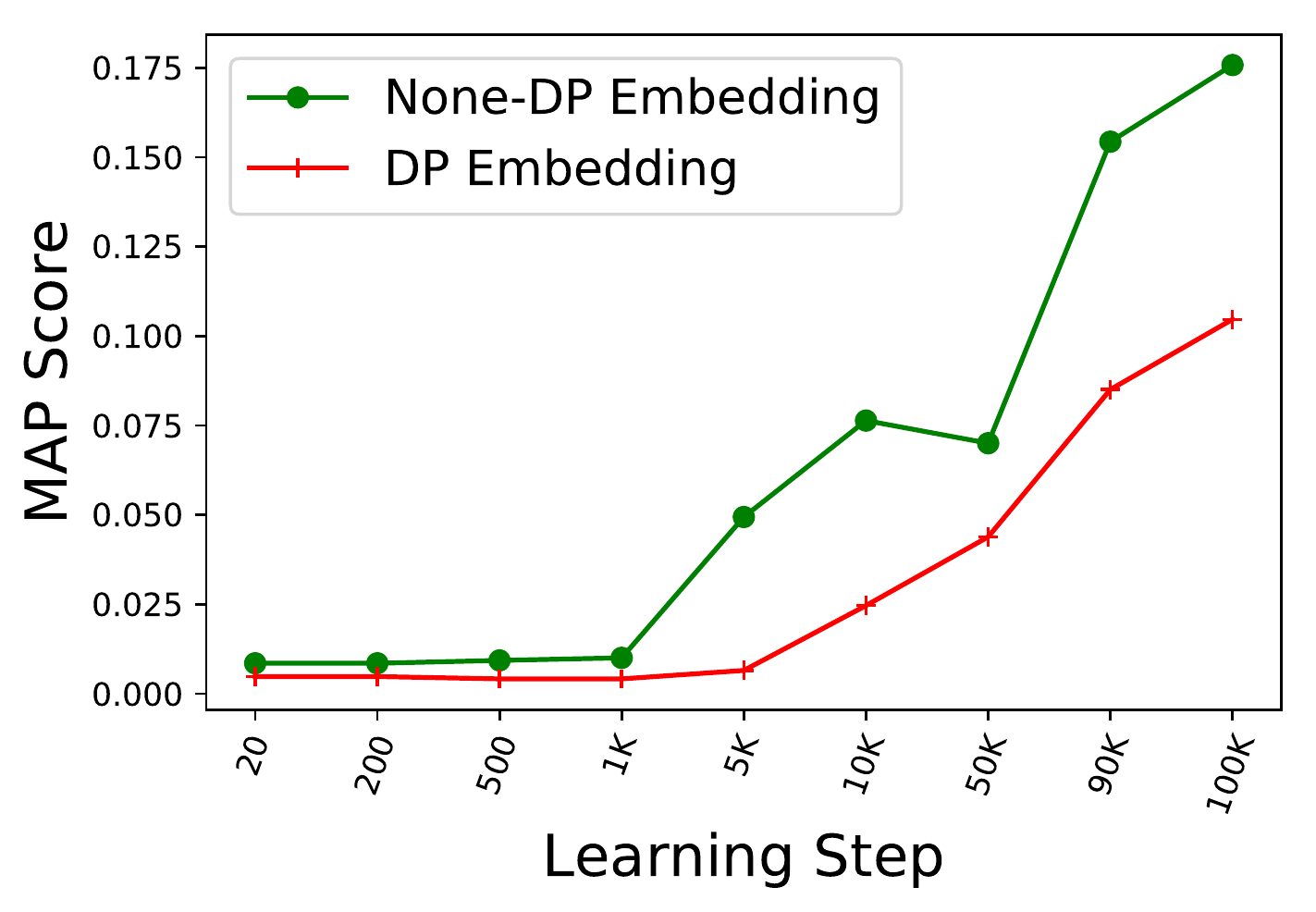}
		\caption{MAP at word level}
		\label{fig:sfig1}
	\end{subfigure}%
	\begin{subfigure}{.5\textwidth}
		\centering
		\includegraphics[width=.99\linewidth]{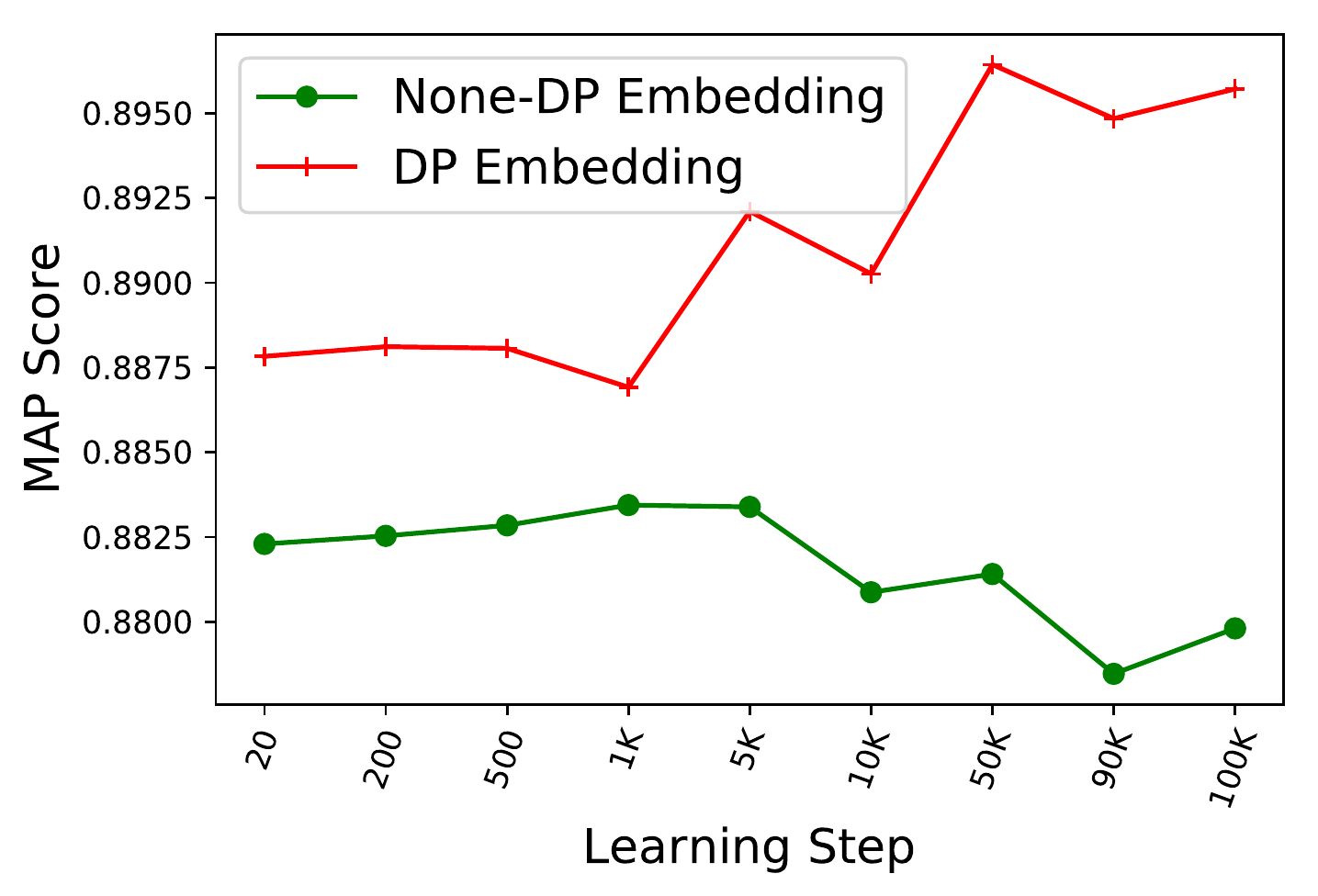}
		\caption{MAP at character level}
		\label{fig:sfig2}
	\end{subfigure}
    \vspace{-0.3cm}
	\caption{Semantic space changes when learning embedding model with and without differential privacy compared to the \emph{Gold model}. Learning step is number of minibatch steps}
	\label{fig:semantic_spaces}
    \vspace{-0.8cm}
\end{figure}

Figure~\ref{fig:semantic_spaces} and Table~\ref{tbl:top_sim_words} show the evaluation on DP-Embedding and Non-DP Embedding regarding semantic space change. Given 11 word samples as input queries, we obtain the top 100 returned words for them from both models, and compare them with the results from \emph{gold model} by calculating $MAP$ score ($K$ = 100, $Q$ = 11).


As shown in Figure~\ref{fig:semantic_spaces}, there is significant difference between DP-Embedding and Non-DP Embedding when comparing to the \emph{Gold model}. In Figure~\ref{fig:semantic_spaces}-(a), it clearly shows that DP-Embedding performs slightly lower performance compared to None-DP Embedding. This is understandable due to the injected noise into the model for differential privacy. However,  one interesting fact we can observe in Figure~\ref{fig:semantic_spaces}-(a) is that even the performance at word-level (using the $MAP$-Word metric) is lower, at character-level (using the $MAP$-Char metric), DP-Embedding performs better than None-DP Embedding. This observation gives a hint that the reasonable noise we inject into the model actually helps the model to improve at character level. Intuitively, injecting noise is similar to modifying characters of words. It is worth to notice that, up to date, this observation is very new and has not reported in any work before. Thus, further verification is worthwhile in future.
Table~\ref{tbl:top_sim_words} presents an example with top 4 results from both models given three queries (``three'', ``eight'', and ``they''). Two types of $MAP$ scores ($MAP$-Word and $MAP$-Char) are calculated. As shown in the third column, some irrelevant concepts (e.g., ``feeder'', or ``stockholders'') are being mixed up at the top in DP-Embedding, while for the None-DP Embedding (the fifth column), the relevant concepts (i.e., ``four'', ``one'', ``it'') are climbing up to the top.  Though relevant concepts are always expected to get closer over each training step for word embedding model, the added noise to the DP-Embedding model over a learning step can create the distance between sensitive concepts further.  In this way, privacy is guaranteed for word embedding model. However, too much noise might destroy the model quality, thus we will evaluate data utility.

\subsection{Evaluation \#2: Regression task} \label{subsec:regression}
As explained in Section \ref{subsec:experimentdesign}, we address the regression problem in this experiment by using DP-Embedding model and None-DP Embedding model respectively. RMSE (root mean square error) is used to evaluate the regression task. Lower RMSE proves better regression performance. Table~\ref{tbl:linear_regression_results} shows that the usage of DP-Embedding gets better or slightly different results than the None-DP Embedding. This clearly shows that, with $(\epsilon, \delta)$ privacy guarantee at some settings, such as (0.125,0.0184)-DP and (0.125,0.0197)-DP, we achieve the optimized trade-off of privacy-guarantee and data utilities. 

\section{Conclusions}
\label{conclusions}
In this work, we proposed algorithms for learning differentially private text representation (i.e., word embeddings) for user generated contents (UGC) sharing.  We empirically evaluated the algorithms on a realistic UGC dataset and demonstrated that the proposed embedding model benefits from sensitive data while maintaining user-privacy. The differentially private word embedding allows information from sensitive data to be shared independently. Differently from the previous works on differential privacy, which simply preserve privacy against adversaries on sensitive data, we trained word embedding model on potentially sensitive user generated contents, and our trained model is applied for data sharing with privacy-guarantee.  As the very first work on publicly shared embedding models, this work highlights the new direction of publicly shared embedding models on sensitive text data.  Much future work remains. 
For example, one promising direction would be exploring strategies to detect what are exactly sensitive contents to certain users (e.g., building a knowledge base) to apply our proposed personalized-privacy guarantee.

\section*{Acknowledgement}
This work is supported by the Federated Database project funded by Ume{\aa} University, Sweden. The computations were performed on resources provided by the Swedish National Infrastructure for Computing (SNIC) at HPC2N center. The authors also thank the myPersonality project for data contribution. 

\bibliographystyle{splncs04}
\bibliography{xuansonResearch.bib}

\begin{thebibliography}{10}
\providecommand{\url}[1]{\texttt{#1}}
\providecommand{\urlprefix}{URL }
\providecommand{\doi}[1]{https://doi.org/#1}

\bibitem{Abadi:2016}
{Abadi}, M., {Chu}, A., {Goodfellow}, I., {Brendan McMahan}, H., {Mironov}, I.,
  {Talwar}, K., {Zhang}, L.: {Deep Learning with Differential Privacy}. ArXiv
  e-prints  (Jul 2016)

\bibitem{Bansal:2014}
Bansal, M., Gimpel, K., Livescu, K.: Tailoring continuous word representations
  for dependency parsing. In: Proceedings of the 52nd Annual Meeting of the
  Association for Computational Linguistics (Volume 2: Short Papers). pp.
  809--815. Association for Computational Linguistics (2014).
  \doi{10.3115/v1/P14-2131}, \url{http://aclweb.org/anthology/P14-2131}

\bibitem{Batmanghelich:2016}
Batmanghelich, K.N., Saeedi, A., Narasimhan, K., Gershman, S.: Nonparametric
  spherical topic modeling with word embeddings. Proceedings of the 54th Annual
  Meeting of the Association for Computational Linguistics (Volume 2: Short
  Papers)  \textbf{abs/1604.00126},  537--542 (2016),
  \url{http://arxiv.org/abs/1604.00126}

\bibitem{Bayardo:2005}
Bayardo, R.J., Agrawal, R.: Data privacy through optimal k-anonymization. In:
  ICDE. pp. 217--228 (2005)

\bibitem{bojanowski2017enriching}
Bojanowski, P., Grave, E., Joulin, A., Mikolov, T.: Enriching word vectors with
  subword information. Transactions of the Association for Computational
  Linguistics  \textbf{5},  135--146 (2017)

\bibitem{Dwork:2006}
Cynthia, D.: Differential privacy. pp. 1--12. ICALP (2006)

\bibitem{Dwork:2009}
Dwork, C., Smithy, A.: Differential privacy for statistics:what we know and
  what we want to learn  (2009)

\bibitem{Ebadi:2015}
Ebadi, H., Sands, D., Schneider, G.: Differential privacy: Now it's getting
  personal. In: Proceedings of the 42Nd Annual ACM SIGPLAN-SIGACT Symposium on
  Principles of Programming Languages. pp. 69--81. POPL '15, ACM, New York, NY,
  USA (2015). \doi{10.1145/2676726.2677005},
  \url{http://doi.acm.org/10.1145/2676726.2677005}

\bibitem{FlekovaG13}
Flekova, L., Gurevych, I.: Can we hide in the web? large scale simultaneous age
  and gender author profiling in social media notebook for {PAN} at {CLEF}
  2013. In: Working Notes for {CLEF} 2013 Conference , Valencia, Spain,
  September 23-26, 2013. (2013)

\bibitem{Fredrikson:2015}
Fredrikson, M., Jha, S., Ristenpart, T.: Model inversion attacks that exploit
  confidence information and basic countermeasures. In: Proceedings of the 22Nd
  ACM SIGSAC Conference on Computer and Communications Security. pp.
  1322--1333. CCS '15 (2015)

\bibitem{Goodfellow:2016}
Goodfellow, I., Bengio, Y., Courville, A.: Deep Learning. MIT Press (2016),
  \url{http://www.deeplearningbook.org}

\bibitem{mypersonality:2015}
Kosinski, M., Matz, S., Gosling, S., Popov, V., Stillwell, D.: Facebook as a
  social science research tool. American Psychologist  (2015)

\bibitem{Lee:2011}
Lee, J., Clifton, C.: How much is enough? choosing $\epsilon$ for differential
  privacy pp. 325--340 (2011)

\bibitem{Lee:2012}
Lee, J., Clifton, C.: Differential identifiability. In: Proceedings of KDD
  (2012)

\bibitem{Levy:2014}
Levy, O., Goldberg, Y.: Linguistic regularities in sparse and explicit word
  representations. In: Proceedings of the Eighteenth Conference on
  Computational Natural Language Learning. pp. 171--180. Association for
  Computational Linguistics (2014). \doi{10.3115/v1/W14-1618},
  \url{http://aclweb.org/anthology/W14-1618}

\bibitem{McMahan:2018}
McMahan, H.B., Ramage, D., Talwar, K., Zhang, L.: Learning differentially
  private language models without losing accuracy. CoRR
  \textbf{abs/1710.06963} (2017), \url{http://arxiv.org/abs/1710.06963}

\bibitem{PINQ:09}
McSherry, F.D.: Privacy integrated queries: An extensible platform for
  privacy-preserving data analysis. In: SIGMOD (2009)

\bibitem{Mikolov:13}
Mikolov, T., Chen, K., Corrado, G., Dean, J.: Efficient estimation of word
  representations in vector space. CoRR  \textbf{abs/1301.3781} (2013),
  \url{http://arxiv.org/abs/1301.3781}

\bibitem{Mohsen:2016}
Mohsen, A.M., El-Makky, N.M., Ghanem, N.: Author identification using deep
  learning. In: 2016 15th IEEE International Conference on Machine Learning and
  Applications (ICMLA). pp. 898--903 (Dec 2016). \doi{10.1109/ICMLA.2016.0161}

\bibitem{Nguyen:2016}
Nguyen, H.H., Imine, A., Rusinowitch, M.: Detecting communities under
  differential privacy. In: Proceedings of the 2016 ACM on Workshop on Privacy
  in the Electronic Society. pp. 83--93. WPES '16, ACM, New York, NY, USA
  (2016). \doi{10.1145/2994620.2994624},
  \url{http://doi.acm.org/10.1145/2994620.2994624}

\bibitem{Papernot:2018}
{Papernot}, N., {Song}, S., {Mironov}, I., {Raghunathan}, A., {Talwar}, K.,
  {Erlingsson}, {\'U}.: {Scalable Private Learning with PATE}. Sixth
  International Conference on Learning Representation (ICLR 2018)  (Feb 2018)

\bibitem{pennington2014glove}
Pennington, J., Socher, R., Manning, C.D.: Glove: Global vectors for word
  representation. In: Empirical Methods in Natural Language Processing (EMNLP).
  pp. 1532--1543 (2014), \url{http://www.aclweb.org/anthology/D14-1162}

\bibitem{Phan:2017}
Phan, N., Wu, X., Hu, H., Dou, D.: Adaptive laplace mechanism: Differential
  privacy preservation in deep learning. CoRR  \textbf{abs/1709.05750} (2017),
  \url{http://arxiv.org/abs/1709.05750}

\bibitem{Popov:2018}
Popov, V., Kudinov, M., Piontkovskaya, I., Vytovtov, P., Nevidomsky, A.:
  Distributed fine-tuning of language models on private data. In: International
  Conference on Learning Representations (2018),
  \url{https://openreview.net/forum?id=HkgNdt26Z}

\bibitem{Taddy:2015}
Taddy, M.: Document classification by inversion of distributed language
  representations. In: Proceedings of the 53rd Annual Meeting of the ACL and
  the 7th International Joint Conference on Natural Language Processing (Volume
  2: Short Papers). pp. 45--49. Association for Computational Linguistics
  (2015). \doi{10.3115/v1/P15-2008}, \url{http://aclweb.org/anthology/P15-2008}

\bibitem{Vu:2018b}
Vu, X.S., Jiang, L.: Self-adaptive privacy concern detection for user-generated
  content. In: Proceedings of the 19th International Conference on
  Computational Linguistics and Intelligent Text Processing (CICLing), Vol.
  Volume 1: Long papers, p., March 2018 (2018)

\bibitem{Vu:2017}
Vu, X.S., Jiang, L., Br\"{a}ndstr\"{o}m, A., Elmroth, E.: Personality-based
  knowledge extraction for privacy-preserving data analysis. In: Proceedings of
  the Knowledge Capture Conference. pp. 45:1--45:4. K-CAP 2017, ACM, New York,
  NY, USA (2017). \doi{10.1145/3148011.3154479},
  \url{http://doi.acm.org/10.1145/3148011.3154479}

\bibitem{WANG:2009}
Wang, R., Wang, X., Li, Z., Tang, H., Reiter, M.K., Dong, Z.:
  Privacy-preserving genomic computation through program specialization. pp.
  338--347. CCS (2009)

\bibitem{Wu:2018}
Wu, Z., Wang, Z., Wang, Z., Jin, H.: Towards privacy-preserving visual
  recognition via adversarial training: A pilot study. In: Ferrari, V., Hebert,
  M., Sminchisescu, C., Weiss, Y. (eds.) Computer Vision -- ECCV 2018. pp.
  627--645. Springer International Publishing, Cham (2018)

\bibitem{Zhang:2018}
Zhang, Y., Ding, N., Soricut, R.: {SHAPED:} shared-private encoder-decoder for
  text style adaptation. The 16th Annual Conference of the North American
  Chapter of the Association for Computational Linguistics: Human Language
  Technologies (NAACL)  (2018)

\end{thebibliography}

\end{document}